\documentclass{acm_proc_article-sp}
\usepackage{ifthen,stmaryrd,color}
\usepackage[utf8]{inputenc}
\usepackage{proof}
\usepackage{acmcopyright}

\newcommand\nl[1]{``\emph{#1}''}
\newcommand\wl[1]{\texttt{#1}}
\newcommand\citep{\cite}
\newcommand\citet{\cite}

\newcommand\myparagraph[1]{\textbf{#1.}}

\newcommand\den[1]{\llbracket #1 \rrbracket_c}
\newcommand\deni[1]{\llbracket #1 \rrbracket_{c\I}}

\newcommand\primes{\wl{primes}}
\newcommand\LessThanTen{(-\infty, 10)}

\newcommand\cat[1]{\text{\sc{#1}}}
\newcommand\N{\cat{N}}
\newcommand\Rel{\cat{N|N}}
\newcommand\NP{\cat{NP}}
\newcommand\QP{\cat{QP}}
\newcommand\ROOT{\cat{ROOT}}
\newcommand\Span[2]{[#1\!:\!#2]}
\newcommand\Length{\text{length}}
\newcommand\Score{\text{score}}

\newcommand\I{_i}

\newcommand\mc[2]{\multicolumn{#1}{c}{#2}}
\newcommand{\derivline}[2]{\mc{#1}{\hrulefill \text{\tiny (#2)} \hrulefill}}
\newcommand{\derivlinez}[1]{\mc{#1}{\hrulefill}}

\newcommand\new[1]{#1}
\newcommand\newer[1]{#1}

\newcommand\sO{\ensuremath{\mathcal{O}}}

\newcommand\FigTop[4]{\begin{figure}[t] \begin{center} \includegraphics[scale=#2]{#1} \end{center} \caption{\label{fig:#3} #4} \end{figure}}

\newcommand\R{\ensuremath{\mathbb{R}}} %
\newcommand\eqdef{\ensuremath{\stackrel{\rm def}{=}}} %
\newcommand{\1}{\mathbb{I}} %
\newcommand\refeqn[1]{(\ref{eqn:#1})}

\newcommand\refsec[1]{Section~\ref{sec:#1}}

\newcommand\reffig[1]{Figure~\ref{fig:#1}}

\ifthenelse{\isundefined{\definition}}{\newtheorem{definition}{Definition}}{}
\ifthenelse{\isundefined{\assumption}}{\newtheorem{assumption}{Assumption}}{}
\ifthenelse{\isundefined{\hypothesis}}{\newtheorem{hypothesis}{Hypothesis}}{}
\ifthenelse{\isundefined{\proposition}}{\newtheorem{proposition}{Proposition}}{}
\ifthenelse{\isundefined{\theorem}}{\newtheorem{theorem}{Theorem}}{}
\ifthenelse{\isundefined{\lemma}}{\newtheorem{lemma}{Lemma}}{}
\ifthenelse{\isundefined{\corollary}}{\newtheorem{corollary}{Corollary}}{}
\ifthenelse{\isundefined{\alg}}{\newtheorem{alg}{Algorithm}}{}
\ifthenelse{\isundefined{\example}}{\newtheorem{example}{Example}}{}
\setcopyright{acmlicensed}

\begin{document}

\title{Learning Executable Semantic Parsers for \\ Natural Language Understanding}

\numberofauthors{1} %
\author{
\alignauthor
Percy Liang \\
\affaddr{Computer Science Department}\\
\affaddr{Stanford University}\\
\affaddr{Stanford, CA}\\
\email{pliang@cs.stanford.edu}
}

\maketitle
\begin{abstract}
  For building question answering systems and natural language interfaces,
semantic parsing has emerged as an important and powerful paradigm.
Semantic parsers map natural language into logical forms,
the classic representation for many important linguistic phenomena.
The modern twist is that we are interested in learning semantic parsers from data,
which introduces a new layer of statistical and computational issues.
This article lays out the components of a statistical semantic parser,
highlighting the key challenges.
We will see that semantic parsing is a rich fusion of the logical and the statistical world,
and that this fusion will play an integral role
in the future of natural language understanding systems.

\end{abstract}


\section{Introduction}

A long-standing goal of artificial intelligence (AI) is to build systems capable of understanding natural language.
To focus the notion of ``understanding'' a bit,
let us say that the system must produce an appropriate action
upon receiving an input utterance from a human.
For example:

\newcommand\inputStr{Utterance}
\newcommand\contextStr{Context}
\newcommand\outputStr{Action}
\begin{tabular}{rl}
  \contextStr: & knowledge of mathematics \\
  \inputStr: & \emph{What is the largest prime less than 10?} \\
  \outputStr: & {7}
\end{tabular}
\vspace{0.05in}
\\
\begin{tabular}{rl}
  \contextStr: & knowledge of geography \\
  \inputStr: & \emph{What is the tallest mountain in Europe?} \\
  \outputStr: & {Mt. Elbrus}
\end{tabular}
\vspace{0.05in}
\\
\begin{tabular}{rl}
  \contextStr: & user's calendar \\
  \inputStr: & \emph{Cancel all my meetings after 4pm tomorrow.} \\
  \outputStr: & (removes meetings from calendar)
\end{tabular}

We are interested in utterances such as the ones above,
which require deep understanding and reasoning.
This article focuses on \emph{semantic parsing},
an area within the field of natural language processing (NLP),
which has been growing over the last decade.
Semantic parsers map input utterances into semantic
representations called \emph{logical forms} that support this form of reasoning.
For example, the first utterance above would map onto the logical form
$\max(\primes \cap \LessThanTen)$.
We can think of the logical form as a program that is \emph{executed}
to yield the desired behavior (e.g., answering $7$).
The second utterance would map onto a database query;
the third, onto an invocation of a calendar API.

Semantic parsing is rooted in formal semantics,
pioneered by logician Richard Montague \citep{montague73ptq},
who famously argued that there is ``no important theoretical difference between
natural languages and the artificial languages of logicians.''
Semantic parsing, by residing in the practical realm, is more exposed
to the differences between natural language and logic,
but \new{it inherits two general insights from formal semantics}:
The first idea is \emph{model theory}, which states that expressions (e.g., $\primes$)
are mere symbols which only obtain their meaning or denotation (e.g., $\{2, 3, 5, \dots\}$)
by executing the expression with respect to a model, or in our terminology, a context.
This property allows us to factor out the understanding of language (semantic parsing)
from world knowledge (execution).
Indeed, one can understand the utterance \nl{What is the largest prime less than 10?}
without actually computing the answer.
The second idea is \emph{compositionality}, a principle often attributed to
Gottlob Frege, which states that the denotation of an expression is defined
recursively in terms of the denotations of its subexpressions.
For example, $\primes$ denotes the set of primes,
$\LessThanTen$ denotes the set of numbers smaller than $10$,
and so $\primes \cap \LessThanTen$ denotes the intersection of those two sets.
This compositionality is what allows us to have a succinct characterization of meaning
for a combinatorial range of possible utterances.

\myparagraph{Early systems}
Logical forms have played a foundational role in natural language
understanding systems since their genesis in the 1960s.
Early examples included
LUNAR, a natural language interface into a database about moon rocks \citep{woods72lunar},
and SHRDLU, a system that could both answer questions and perform actions
in a toy blocks world environment \citep{winograd1972language}.
For their time, these systems were significant achievements.
They were able to handle fairly complex linguistic phenomena and integrate
syntax, semantics, and reasoning in an end-to-end application.
For example, SHRDLU was able to process
\nl{Find a block which is taller than the one you are holding and put it into the box.}
However, as the systems were based on hand-crafted rules,
it became increasingly difficult to generalize beyond the narrow domains and handle
the intricacies of general language.
\myparagraph{Rise of machine learning}
In the early 1990s, influenced by the successes of statistical techniques
in the neighboring speech recognition community,
the field of NLP underwent a statistical revolution.
Machine learning offered a new paradigm:
Collect \emph{examples} of the desired input-output behavior
and then fit a statistical model to these examples.
The simplicity of this paradigm coupled with the increase in data and computation 
allowed machine learning to prevail.

What fell out of favor was not only rule-based methods,
but also the natural language understanding problems.
In the statistical NLP era, much of the community's attention
turned to tasks---documentation classification, part-of-speech tagging, and
syntactic parsing---which fell short of full end-to-end understanding.
Even question answering systems relied less on understanding
and more on a shallower analysis coupled with a large collection of
unstructured text documents \citep{brill2002askmsr},
typified by the TREC competitions.
\myparagraph{Statistical semantic parsing}
The spirit of deep understanding was kept alive
by researchers in statistical semantic parsing
\citep{zelle96geoquery,miller96statistical,wong07synchronous,zettlemoyer05ccg,kwiatkowski10ccg}.
A variety of different semantic representations and learning algorithms were employed,
but all of these approaches relied on having a labeled dataset of natural language utterances
paired with annotated logical forms, for example:

\begin{tabular}{rl}
  Utterance: & \emph{What is the largest prime less than 10?} \\
  Logical form: & $\max(\primes \cap \LessThanTen)$
\end{tabular}

\myparagraph{Weak supervision}
Over the last few years, two exciting developments have really spurred interest
in semantic parsing.
The first is reducing the amount of supervision from annotated logical forms
to answers \citep{clarke10world,liang11dcs}: %

\begin{tabular}{rl}
  Utterance: & \emph{What is the largest prime less than 10?} \\
  Action: & 7
\end{tabular}

This form of supervision is much easier to obtain via crowdsourcing.
Although the logical forms are not observed, they are still modeled as
latent variables, which must be inferred from the answer.
This results in a more difficult learning problem,
but \citet{liang11dcs} showed that it is possible to solve it without degrading accuracy.

\myparagraph{Scaling up}
The second development is the scaling up of semantic parsers to more complex domains.
Previous semantic parsers had only been trained on limited domains such as US geography,
but the creation of broad-coverage knowledge bases
such as Freebase \citep{bollacker2008freebase}
set the stage for a new generation
of semantic parsers for question answering.
Initial systems %
required annotated logical forms \citep{cai2013large},
but soon, systems became trainable from answers
\citep{berant2013freebase,kwiatkowski2013scaling,berant2014paraphrasing}.
Semantic parsers have even been extended beyond fixed knowledge bases
to semi-structured tables \citep{pasupat2015compositional}.
With the ability to learn semantic parsers from question-answer pairs,
it is easy to collect datasets via crowdsourcing.
As a result, semantic parsing datasets have grown by an order of magnitude.

In addition, semantic parsers have been applied to a number of applications
outside question answering:
robot navigation \citep{tellex2011understanding,artzi2013weakly},
identifying objects in a scene \citep{matuszek2012grounded,krishnamurthy2013jointly},
converting natural language to regular expressions \citep{kushman2013regex},
and many others.
\myparagraph{Outlook}
Today, the semantic parsing community is a vibrant field,
but it is still young and grappling with the complexities of the natural
language understanding problem.
Semantic parsing poses three types of challenges: %
\begin{itemize}
  \setlength\itemsep{0pt}
  \item Linguistic:
    How should we represent the semantics of natural language and construct it compositionally from
    the natural language?
  \item Statistical:
    How can we learn semantic parsers from weak supervision
    and \emph{generalize} well to new examples?
  \item Computational: How do we efficiently search over the combinatorially
    large space of possible logical forms?
\end{itemize}

The rest of the paper is structured as follows:
We first present a general framework for semantic parsing,
introducing the key components (\refsec{framework}).
The framework is pleasantly modular,
with different choices of the components corresponding to
existing semantic parsers in the literature (\refsec{refinement}).
We describe the datasets (\refsec{datasets})
and then conclude (\refsec{discussion}).

\section{Framework}
\label{sec:framework}

\FigTop{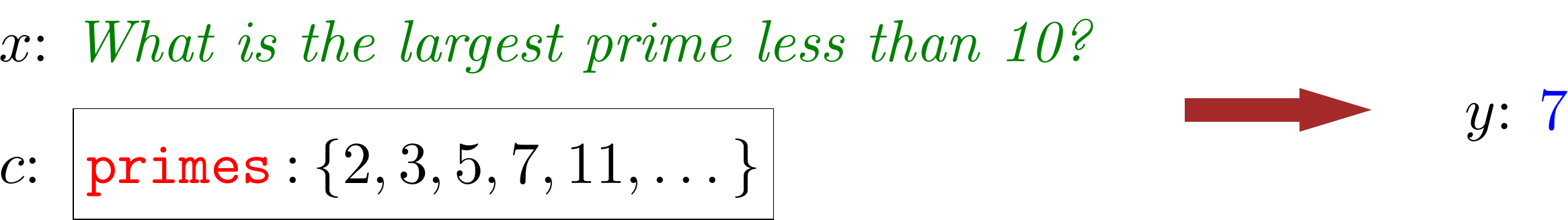}{0.3}{problem}{A natural language understanding problem
where the goal is to map an utterance $x$ in a context $c$ to an action $y$.
}

\myparagraph{Natural language understanding problem}
In this article, we focus on the following natural language understanding
problem:~Given an \emph{utterance} $x$ in a \emph{context} $c$, output the
desired \emph{action} $y$.
\reffig{problem} shows the setup for a question answering application,
in which case $x$ is a question, $c$ is a knowledge base,
and $y$ is the answer.
In a robotics application, $x$ is a command, $c$ represents the robot's environment,
and $y$ is the desired sequence of actions to be carried by the robot
\citep{tellex2011understanding}.
To build such a system, assume that we are given a set of $n$ examples
$\{(x\I,c\I,y\I)\}_{i=1}^n$.
We would like to use these examples to train a model that can generalize
to new unseen utterances and contexts.

\myparagraph{Semantic parsing components}
This article focuses on a statistical semantic parsing approach to the above problem,
where the key is to posit an intermediate \emph{logical form} $z$ that connects $x$ and $y$.
Specifically, $z$ captures the semantics of the utterance $x$,
and it also executes to the action $y$ (in the context of $c$).
In our running example, $z$ would be
$\wl{max}(\primes \cap \LessThanTen)$.
Our semantic parsing framework consists of the following five components
(see \reffig{framework}):
\begin{enumerate}
  \setlength{\itemsep}{0pt}
\item \textbf{Executor}:
  computes the denotation (action) $y = \den{z}$ given a logical form $z$ and context $c$.
  This defines the semantic representation (logical forms along with their denotations).
\item \textbf{Grammar}: a set of rules $G$
  that produces $D(x, c)$, a set of candidate derivations of logical forms.
\item \textbf{Model}: specifies a distribution $p_\theta(d \mid x, c)$
  over derivations $d$ parameterized by $\theta$.
\item \textbf{Parser}: searches for high probability derivations $d$
  under the model $p_\theta$.
\item \textbf{Learner}: estimates the parameters $\theta$ (and possibly rules in $G$)
  given training examples $\{(x\I,c\I,y\I)\}_{i=1}^n$.
\end{enumerate}

\FigTop{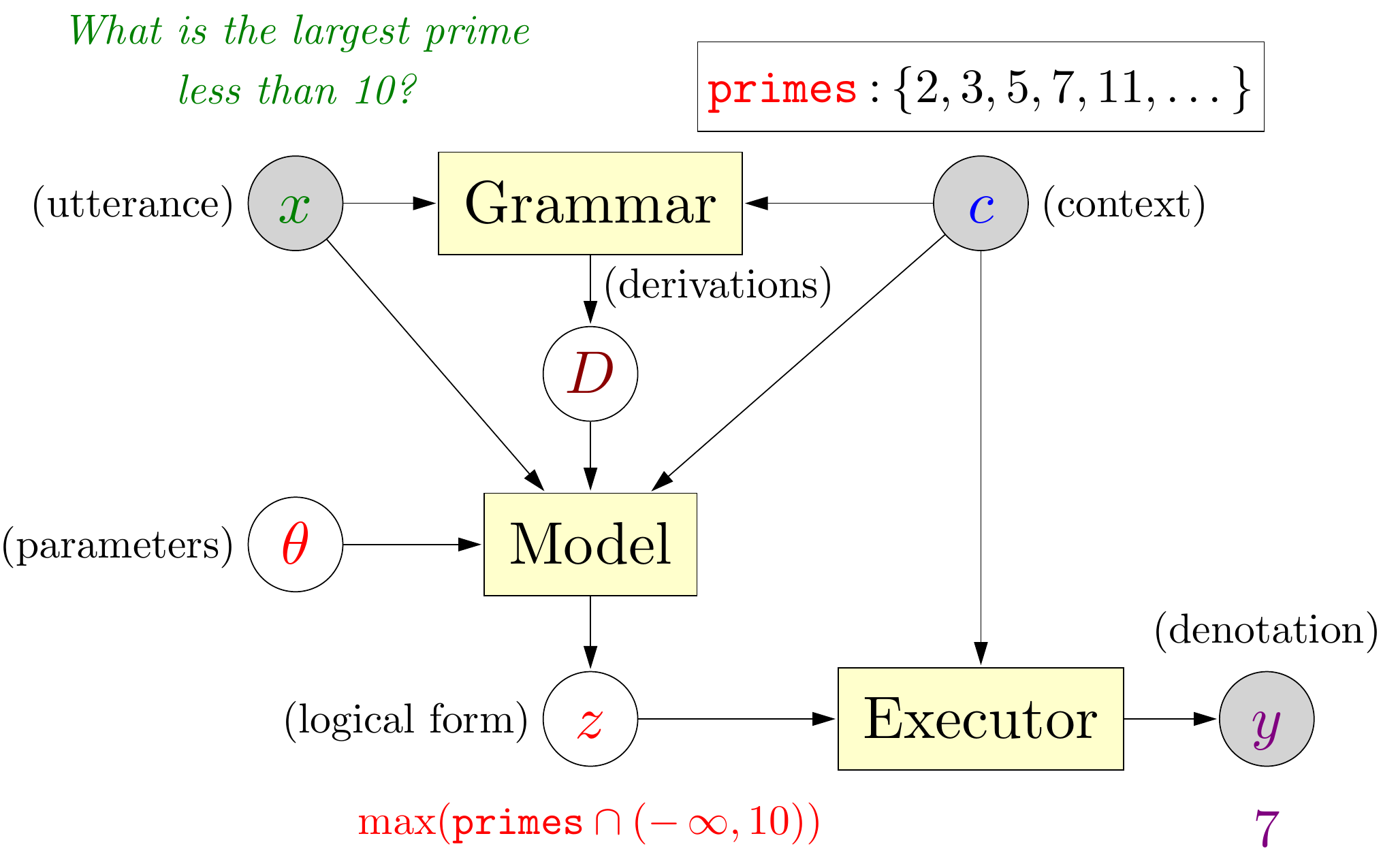}{0.3}{framework}{
  Semantic parsing framework depicting the executor, grammar, and model.
  The parser and learner are algorithmic components that are responsible
  for generating the logical form $z$ and parameters $\theta$, respectively.
}

We now instantiate each of these components for our running example:
\nl{What is the largest prime less than 10?}

\myparagraph{Executor}
Let the semantic representation be the language of mathematics,
and the executor is the standard interpretation,
where the interpretations of predicates (e.g., $\primes$) are given by $c$.
With $c(\primes) = \{ 2, 3, 5, 7, 11, \dots, \}$,
the denotation is $\den{\primes \cap \LessThanTen} = \{ 2, 3, 5, 7 \}$.

\myparagraph{Grammar}
The grammar $G$ connects utterances to possible \emph{derivations} of logical forms.
Formally, the grammar is a set of rules of the form $\alpha \Rightarrow \beta$.\footnote{
  \new{The standard way context-free grammar rules are written is $\beta \to \alpha$.
  Because our rules build logical forms, reversing the arrow is more natural.}
}
Here is a simple grammar for our running example:
\begin{center}
\begin{tabular}{llcl}
  \hline
  (R1) & \emph{prime} & $\Rightarrow$ & $\NP[\primes]$ \\
  (R2) & \emph{10} & $\Rightarrow$ & $\NP[10]$ \\
  (R3) & \emph{less than} $\NP[z]$ & $\Rightarrow$ & $\QP[(-\infty, z)]$ \\
  (R4) & $\NP[z_1]$ $\QP[z_2]$ & $\Rightarrow$ & $\NP[z_1 \cap z_2]$ \\
  (R5) & \emph{largest} $\NP[z]$ & $\Rightarrow$ & $\NP[\max(z)]$ \\
  (R6) & \emph{largest} $\NP[z]$ & $\Rightarrow$ & $\NP[\min(z)]$ \\
  (R7) & \emph{What is the} $\NP[z]$? & $\Rightarrow$ & $\ROOT[z]$ \\
  \hline
\end{tabular}
\end{center}

We start with the input utterance and repeatedly apply rules in $G$.
A rule $\alpha \Rightarrow \beta$ can be applied if some span of the utterance matches $\alpha$,
in which case a derivation over the same span with a new syntactic category
and logical form according to $\beta$ is produced.
Here is one possible derivation (call it $d_1$) for our running example:

\begin{align}
\begin{array}{*{6}{c}}
\emph{What is the} & \emph{largest} & \emph{prime}       & \emph{less than} & \emph{10}          & \emph{?} \\
                   &                & \derivline{1}{R1}  &                  & \derivline{1}{R2}  &          \\
                   &                & \NP[\primes]       &                  & \NP[10]            &          \\
                   &                &                    & \derivline{2}{R3}                     &          \\
                   &                &                    & \mc{2}{\QP[(-\infty, 10)]}            &          \\
                   &                & \derivline{3}{R4}                                          &          \\
                   &                & \mc{3}{\NP[\primes \cap (-\infty, 10)]}                    &          \\
                   & \derivline{4}{R5}                                                           &          \\
                   & \mc{4}{\NP[\max(\primes \cap (-\infty, 10))]}                               &          \\
\derivline{6}{R7}                                                                                           \\
\mc{6}{\ROOT[\max(\primes \cap (-\infty, 10))]}                                                             \\
\end{array}
\end{align}

For example, applying (R3) produces category $\QP$ and logical form $(-\infty, 10)$
over span $\Span{5}{7}$ corresponding to \nl{less than 10}.
We stop when we produce the designated $\ROOT$ category over the entire utterance.
Note that we could have also applied (R6) instead of (R5) to generate
the (incorrect) logical form $\min(\primes \cap \LessThanTen)$;
let this derivation be $d_2$.
We have $D(x, c) = \{ d_1, d_2 \}$ here,
but in general, there could be exponentially many derivations,
and multiple derivations can generate the same logical form.
\new{In general, the grammar might contain nonsense rules (R6)
that do not reflect ambiguity in language
but are rather due to model uncertainty prior to learning.}

\myparagraph{Model}
The model scores the set of candidate derivations generated by the grammar.
A common choice used by virtually all existing semantic parsers
are log-linear models (generalizations of logistic regressions).
In a log-linear model,
define a \emph{feature vector} $\phi(x, c, d) \in \R^F$ for each possible derivation $d$.
We can think of each feature as casting a vote for various derivations $d$ based on some coarse property
of the derivation. %
For example, define $F = 7$ features,
each counting the number of times a given grammar rule is invoked in $d$,
so that $\phi(x,c,d_1) = [1, 1, 1, 1, 1, 0, 1]$ and
$\phi(x,c,d_2) = [1, 1, 1, 1, 0, 1, 1]$.

Next, let $\theta \in \R^F$ denote the \emph{parameter vector},
which defines a weight for each feature representing how reliable that feature is.
Their weighted combination $\Score(x, c, d) = \phi(x,c,d) \cdot \theta$ represents how good the derivation is.
We can exponentiate and normalize these scores to obtain a distribution over derivations:
\begin{align}
  \label{eqn:model}
  p_\theta(d \mid x, c) = \frac{\exp(\Score(x, c, d))}{\sum_{d' \in D(x, c)} \exp(\Score(x, c, d'))}.
\end{align}
If $\theta = [0, 0, 0, 0, +1, -1, 0]$,
then $p_\theta$ would assign probability $\frac{\exp(1)}{\exp(1) + \exp(-1)} \approx 0.88$ to $d_1$
and $\approx 0.12$ to $d_2$.

\myparagraph{Parser}
Given a trained model $p_\theta$, the parser (approximately) computes the
highest probability derivation(s) for an utterance $x$ under $p_\theta$.
Assume the utterance $x$ is represented as a sequence of tokens (words).
A standard approach is to use a \emph{chart parser},
which recursively builds derivations for each span of the utterance.
Specifically, for each category $A$ and span $\Span{i}{j}$ (where $0 \le i < j \le \Length(x)$),
we loop over the applicable rules in the grammar $G$ and apply each one to build
new derivations of category $A$ over $\Span{i}{j}$.
For binary rules---those of the form $B \, C \Rightarrow A$ such as (R4),
we loop over split points $k$ (where $i < k \le j$), recursively compute
derivations $B[z_1]$ over $\Span{i}{k}$ and $C[z_2]$ over $\Span{k}{j}$,
and combine them into a new derivation $A[z]$ over $\Span{i}{j}$,
where $z$ is determined by the rule; for example, $z = z_1 \cap z_2$ for (R4).
The final derivations for the utterance are collected in the $\ROOT$ category over
span $\Span{0}{\Length(x)}$.

The above procedure would generate all derivations,
which could be exponentially large.
Generally, we only wish to compute compute the derivations with high
probability under our model $p_\theta$.
If the features of $p_\theta$ were to \emph{decompose} as a sum over the rule
applications in $d$---that is,
$\phi(x, c, d) = \sum_{(r, i, j) \in d} \phi_\text{rule}(x, c, r, i, j)$,
then we could use dynamic programming:
\newer{For each category $A$ over $\Span{i}{j}$,
compute the highest probability derivation.}
\new{However, in executable semantic parsing,
feature decomposition isn't sufficient, since during learning,
we also need to incorporate the constraint that the logical form executes to the
true denotation ($\1[\den{d.z} = y]$); see \refeqn{update} below.
}
\newer{
To maintain exact computation in this setting,
the dynamic programming state would need to include
the entire logical form $d.z$,
which is infeasible,
since there are exponentially many logical forms.
}
Therefore, \emph{beam search} is generally employed,
where we keep only the $K$ sub-derivations with the highest model score
based on only features of the sub-derivations. %
Beam search is not guaranteed to return the $K$ highest
scoring derivations, but it is often an effective heuristic.

\myparagraph{Learner}
While the parser turns parameters into derivations, the learner solves
the inverse problem. %
The dominant paradigm in machine learning is to set up an objective function
and optimize it. %
A standard principle is to maximize the likelihood of the training data
$\{(x\I, c\I, y\I)\}_{i=1}^n$.
An important point is that we don't observe the correct derivation for
each example, but only the action $y\I$,
so we must consider all derivations $d$
whose logical form $d.z$ satisfy $\deni{d.z} = y\I$.
This results in the log-likelihood of the observed action $y\I$:
\begin{align}
  \sO\I(\theta) &\eqdef \log \sum_{\stackrel{d \in D(x\I, c\I)}{\deni{d.z} = y\I}} p_\theta(d \mid x\I, c\I).
\end{align}
The final objective is then simply the sum across all $n$ training examples:
\begin{align}
  \sO(\theta) &\eqdef \sum_{i=1}^n \sO\I(\theta),
\end{align}
The simplest approach to maximize $\sO(\theta)$ is to use \emph{stochastic gradient descent} (SGD),
an iterative algorithm that takes multiple passes (e.g., say 5) over the training data
and makes the following update on example $i$:
\begin{align}
  \theta \leftarrow \theta + \eta \nabla \sO\I(\theta),
\end{align}
where $\eta$ is a step size that governs how aggressively we want to update parameters (e.g., $\eta = 0.1$).
In the case of log-linear models, the gradient has a nice interpretable form:
\begin{align}
  \label{eqn:update}
  \nabla \sO\I(\theta) = \sum_{d \in D(x\I, c\I)} (q(d) - p_\theta(d \mid x\I, c\I)) \phi(x\I, c\I, d),
\end{align}
where $q(d) \propto p_\theta(d \mid x\I, c\I) \1[\deni{d.z} = y\I]$
is the model distribution $p_\theta$ over derivations $d$,
but restricted to ones consistent with $y\I$.
The gradient pushes $\theta$ to put more probability mass on $q$ and less on $p_\theta$.
For example, if $p_\theta$ assigns probabilities $[0.2, 0.4, 0.1, 0.3]$ to four derivations
and the middle two derivations are consistent,
then $q$ assigns probabilities $[0, 0.8, 0.2, 0]$.

The objective function $\sO(\theta)$ is not concave,
so SGD is at best guaranteed to converge to a local optimum, not a global one.
Another problem is that we cannot enumerate all derivations $D(x\I, c\I)$ generated by the grammar,
so we approximate this set with the result of beam search,
which yields $K$ candidates (typically $K = 200$);
$p_\theta$ is normalized over this set.
Note that this candidate set depends on the current parameters $\theta$,
resulting a heuristic approximation of the gradient $\nabla \sO\I$.

\myparagraph{Summary}
We have covered the components of a semantic parsing system.
Observe that the components are relatively loosely coupled:~The executor
is concerned purely with what we want to express
independent of how it would be expressed in natural language.
The grammar describes how candidate logical forms are constructed from the
utterance but does not provide algorithmic guidance nor specify a way to score the candidates.  
The model focuses on a particular derivation and defines features
that could be helpful for predicting accurately.
The parser and the learner provide algorithms largely independent of semantic representations.
This modularity allows us to improve each component in isolation.

\section{Refining the Components}
\label{sec:refinement}

Having toured the components of a semantic parsing system, we now return to
each component and discuss the key design decisions and possibilities for improvement.

\subsection{Executor}

By describing an executor, we are really describing the language of the logical form.
A basic textbook representation of language is \emph{first-order logic},
which can be used to make quantified statements about relations between objects.
For example, \nl{Every prime greater than two is odd.} would be expressed in
first-order logic as $\forall x . \wl{prime}(x) \wedge \wl{more}(x, 2) \to \wl{odd}(x)$.
Here, the context $c$ is a model (in the model theory sense), which maps
predicates to sets of objects or object pairs.
The execution of the above logical form with respect to the standard mathematical context
would be $\wl{true}$.
\citet{blackburn05semantics} gives a detailed account on how first-order logic
is used for natural language semantics.

First-order logic is reasonably powerful,
but it fails to capture some common phenomena in language.
For example, \nl{How many primes are less than 10?} requires constructing a set
and manipulating it and thus goes beyond the power of first-order logic.
We can instead augment first-order logic with constructs from \emph{lambda calculus}.
The logical form corresponding to the above question
would be $\wl{count}(\lambda x . \wl{prime}(x) \wedge \wl{less}(x, 10))$,
where the $\lambda$ operator can be thought of as constructing
a set of all $x$ that satisfy the condition;
in symbols,
$\den{\lambda x . f(x)} = \{ x : \den{f(x)} = \wl{true} \}$.
Note that $\wl{count}$ is a higher-order functions that takes a function as an argument.

Another logical language, which can be viewed as syntactic sugar for lambda
calculus, is \emph{lambda dependency-based semantics (DCS)}
\citep{liang2013lambdadcs}.  
In lambda DCS, the above logical form would be
$\wl{count}(\wl{prime} \sqcap (\wl{less}.10))$,
where the constant $10$ represent $\lambda x . (x = 10)$,
the intersection operator $z_1 \sqcap z_2$ represents $\lambda x . z_1(x) \wedge z_2(x)$, and
the join operator $r.z$ represents $\lambda x . \exists y . r(x, y) \wedge z(y)$.

Lambda DCS is ``lifted'' in the sense that operations
combine functions from objects to truth values (think sets) rather than truth values.
As a result, lambda DCS logical forms partially eliminate the need for variables.
Noun phrases in natural language (e.g., \nl{prime less than 10}) also denote sets.
Thus, lambda DCS arguably provides
a transparent interface with natural language.
From a linguistic point of view, the logical language seeks primarily to model
natural language phenomena.
From an application point of view, the logical language dictates what actions we support.
It is thus common to use application-specific logical forms, for example,
regular expressions \citep{kushman2013regex}.
\new{
Note that the other components of the framework are largely independent of the exact
logical language used.
}
\subsection{Grammar}

Recall that the goal of the grammar in this article is just to define a set of
candidate derivations for each utterance and context.
Note that this is in contrast to a conventional notion of a grammar in linguistics,
where the goal is to precisely characterize the set of valid sentences and interpretations.
This divergence is due to two factors:
First, we will learn a statistical model over the derivations generated by the grammar anyway,
so the grammar can be simple and coarse.
Second, we might be interested in application-specific logical forms.
In a flight reservation domain,
the logical form we wish to extract from
\nl{I'm in Boston and would like to go to Portland}
is
$\wl{flight} \sqcap \wl{from}.\wl{Boston} \sqcap \wl{to}.\wl{Portland}$,
which is certainly not the full linguistic meaning of the utterance,
but suffices for the task at hand.
Note that the connection here between language and logic is less direct
compared to \nl{prime less than 10} $\Rightarrow \wl{prime} \sqcap (\wl{less}.10)$.

\myparagraph{CCG}
\new{One common approach to the grammar in semantic parsing
is \emph{combinatory categorial grammar} (CCG) \citep{steedman00ccg}},
which had been developed extensively in linguistics
before it was first used for semantic parsing \citep{zettlemoyer05ccg}.
CCG is typically coupled with logical forms in lambda calculus.
Here is an example of a CCG derivation:

\begin{align}
  \small
\begin{array}{*{3}{c}}
\emph{prime}                        & \emph{less than}                                                                      & \emph{10}         \\
\derivlinez{1}                      & \derivlinez{1}                                                                        & \derivlinez{1}    \\
\N[\lambda x . \wl{prime}(x)]       & (\N\backslash\N)/\NP[\lambda y . \lambda f . \lambda x . f(x) \wedge \wl{less}(x, y)  & \NP[10]           \\
                                    & \derivline{2}{>}                                                                                          \\
                                    & \mc{2}{\N\backslash\N[\lambda f . \lambda x . f(x) \wedge \wl{less}(x, 10)]}                              \\
\derivline{3}{<}                                                                                                                                \\
\mc{3}{\N[\lambda x . \wl{prime}(x) \wedge \wl{less}(x, 10)]}                                                                                   \\
\end{array}
\end{align}
Syntactic categories in CCG include nouns ($\N$) denoting sets
and noun phrases ($\NP$) denoting objects.
Composite categories ($(\N\backslash\N)/\NP$) \newer{represent functions of multiple arguments},
but where the directionality of the slashes indicate the location of the arguments.
Most rules in CCG are lexical (e.g., [\emph{prime} $\Rightarrow \N[\lambda x. \wl{prime}(x)$]),
which mention particular words.
The rest of the rules are glue;
for example, we have a forward application ($>$) and backward application ($<$) rule:
\begin{center}
\begin{tabular}{lllcl}
  \hline
  (>) & $A/B[f]$ & $B[x]$ & $\Rightarrow$ & $A[f(x)]$ \\
  (<) & $B[x]$   & $A \backslash B[f]$ & $\Rightarrow$ & $A[f(x)]$ \\
  \hline
\end{tabular}
\end{center}
It is common to use other combinators which both handle more complex linguistic phenomena
such as NP coordination \nl{integers that divide 2 or 3}
as well as ungrammatical language \citep{zettlemoyer07relaxed},
although these issues can also be handled by having a more expansive lexicon \citep{kwiatkowski10ccg}.

\myparagraph{\new{Crude rules}} %
In CCG, the lexical rules can become quite complicated. %
\new{An alternative approach \newer{that appears mostly in work on lambda DCS}
is to adopt a much cruder grammar} whose rules correspond directly to the lambda DCS constructions,
join and intersect.
This results in the following derivation:
\begin{align}
\small
\begin{array}{*{3}{c}}
\emph{prime}                        & \emph{less than}                & \emph{10}         \\
\derivlinez{1}                      & \derivlinez{1}                  & \derivlinez{1}    \\
\N[\wl{prime}]                      & \Rel[\wl{less}]                 & \N[10]            \\
                                    & \derivline{2}{join}                                 \\
                                    & \mc{2}{\N[\wl{less}.10]}                            \\
\derivline{3}{intersect}                                                                  \\
\mc{3}{\N[\wl{prime} \sqcap \wl{less}.10]}                                                \\
\end{array}
\end{align}
\begin{center}
\begin{tabular}{lllcl}
  \hline
  (join)      & $\Rel[r]$ & $\N[z]$ & $\Rightarrow$ & $\N[r.z]$ \\
  (intersect) & $\N[z_1]$ & $\N[z_2]$ & $\Rightarrow$ & $\N[z_1 \sqcap z_2]$ \\
  \hline
\end{tabular}
\end{center}
The lack of constraints permits the derivation of other logical forms
such as $\wl{10} \sqcap \wl{less}.\wl{prime}$,
but these incorrect logical forms can be ruled out statistically via the model.
The principle is to keep the lexicon simple and \new{lean
more heavily on features (whose weights are learned from data) to derive
drive the selection of logical forms.}

\myparagraph{Floating rules}
While this new lexicon is simpler, the bulk of the complexity comes from having
it in the first place. Where does the rules
[\emph{prime} $\Rightarrow \Rel: \wl{prime}$] and [\emph{less than} $\Rightarrow \Rel: \wl{less}$] come from?  
We can reduce the lexicon even further by introducing \emph{floating} rules
[$\emptyset \Rightarrow \Rel: \wl{prime}$] and [$\emptyset \Rightarrow \Rel: \wl{less}$],
which can be applied in the absence of any lexical trigger.
This way, the utterance \nl{primes smaller than 10} would also be able to
generate $\wl{prime} \sqcap \wl{less}.10$,
where the predicates $\wl{prime}$ and $\wl{less}$
are not anchored to any specific words.

While this relaxation may seem linguistically blasphemous,
recall that the purpose of the grammar is to merely deliver a set of logical forms,
so floating rules are quite sensible provided we can keep the set of logical forms under control.
Of course, since the grammar is so flexible,
even more nonsensical logical forms are generated,
so we must lean heavily on the features to score the derivations properly.
When the logical forms are simple and we have strong type constraints,
this strategy can be quite effective
\citep{berant2014paraphrasing,wang2015overnight,pasupat2015compositional}.

The choice of grammar is arguably the most important component of a semantic parser,
as it most directly governs the tradeoff between expressivity
and statistical/computational complexity.
We have explored three grammars, from a tightly-constrained CCG to
a very laissez faire floating grammar.
CCG is natural for capturing complex linguistic phenomena in well-structured sentences,
whereas for applications where the utterances are noisy but the logical forms are simple,
more flexible grammars are appropriate.
In practice, one would probably want to mix and match
depending on the domain.

\subsection{Model}
\label{sec:model}

At the core of a statistical model is a function $\Score(x, c, d)$
that judges how good a derivation $d$ is with respect to the utterance $x$ and context $c$.
In \refsec{framework}, we described a simple log-linear model
\refeqn{model} in which $\Score(x, c, d) = \phi(x, c, d) \cdot \theta$,
and with simple rule features.
There are two ways to improve the model:
use more expressive features and
and use a non-linear scoring function.

Most existing semantic parsers stick with a linear model and leverage more targeted features.
One simple class of features $\{\phi_{a,b}\}$ is equal to the
number of times word $a$ appears in the utterance and predicate $b$
occurs in the logical form;
if the predicate $b$ is generated by a floating rule,
then $a$ is allowed to range over the entire utterance;
otherwise, it must appear in the span of $b$.

The above features are quite numerous, which provides flexibility
but require a lot of data.
Note that logical predicates (e.g., \wl{birthplace}) and the natural language
(e.g., \emph{born in}) often share a common vocabulary (English words).
One can leverage this structure by defining a few features
based on statistics from large corpora, external lexical resources, or simply
string overlap.
For example, $\phi_\text{match}$ might be the number of word-predicate pairs
with that match almost exactly.
This way, the lexical mapping need not be learned
from scratch using only the training data
\citep{berant2013freebase,kwiatkowski2013scaling}.

A second class of useful features are based on the denotation $\den{d.z}$ of the
predicted logical form.  For example, in a robot control setting, a feature would
encode whether the predicted actions are valid in the environment $c$
(picking up a cup requires the cup to be present).
These features make it amply clear that semantic parsing is not simply
a natural language problem but one that involves jointly reasoning
about the non-linguistic world.
In addition, we typically include type constraints (features with weight $-\infty$)
to prune out ill-formed logical forms such as $\wl{birthplace}.4$
that are licensed the grammar.

One could also adopt a non-linear scoring function,
which does not require any domain knowledge about semantic parsing.
For example, one could use a simple one-layer neural network,
which takes a weighted combination of $m$ non-linear basis functions: %
$\Score(x, c, d) = \sum_{i=1}^m \alpha_i \tanh(\phi(x, c, d) \cdot w_i)$.
\new{The parameters of the model that we would learn are then the $m$ top-level weights $\{ \alpha_i \}$
and the $m F$ bottom-level weights $\{ w_{ij} \}$;
recall that $F$ is the number of features ($\phi(x,c,d) \in \R^F$).}
With more parameters, the model becomes more powerful,
but also requires more data to learn.

\subsection{Parsing}

Most semantic parsing algorithms are based on chart parsing,
where we recursively generate a set of candidate derivations for each syntactic category (e.g., $\NP$)
and span (e.g., $\Span{3}{5}$).
There are two disadvantages of chart parsing:
First, it does not easily support incremental contextual interpretation:
The features of a span $\Span{i}{j}$ can only depend on the sub-derivations in that span,
not on the derivations constructed before $i$.
This makes anaphora (resolution of \nl{it}) difficult to model.
A solution is to use shift-reduce parsing rather than chart parsing \citep{zhao2015type}.
Here, one parses the utterance from left to right, and new sub-derivations
can depend arbitrarily on the sub-derivations constructed thus far.

A second problem is that the chart parsing generally builds derivations
in some fixed order---e.g., of increasing span size.
This causes the parser to waste equal resources on non-promising spans which are unlikely
to participate in the final derivation that is chosen.
This motivates agenda-based parsing, in which derivations that are deemed more promising
by the model are built first \citep{berant2015agenda}.

\subsection{Learner}

In learning, we are given examples of utterance-context-response triples $(x,c,y)$.
There are two aspects of learning: inducing the grammar rules
and estimating the model parameters.
It is important to remember that practical semantic parsers do not do everything from scratch,
and often the hard-coded grammar rules are as important as the training examples.
First, some lexical rules that map named entities (e.g., [\emph{paris} $\Rightarrow \wl{ParisFrance}$]),
dates, and numbers are generally assumed to be given \new{\citep{zettlemoyer05ccg}},
though we need not assume that these rules are perfect \new{\citep{liang11dcs}}.
These rules are also often represented implicitly \new{\citep{liang11dcs,berant2013freebase}}.

How the rest of the grammar is handled varies across approaches.
In CCG-style approach, inducing \new{lexical} rules is an important part of learning.
In \citet{zettlemoyer05ccg}, a procedure called GENLEX is used to generate candidate
lexical rules from a utterance-logical form pair $(x,z)$.
A more generic induction algorithm based on higher-order unification
does not require any initial grammar \citep{kwiatkowski10ccg}.
\new{\citep{wong07synchronous} use machine translation
ideas to induce a synchronous grammar
(which can also be used to generate utterances from logical forms).}
However, all these lexicon induction methods require annotated logical forms $z$.
In approaches that learn from denotations $y$ \citep{liang11dcs,berant2013freebase},
an initial crude grammar is used to generate candidate logical forms,
and rest of the work is done by the features.

As we discussed earlier, parameter estimation can be performed by stochastic
gradient descent on the log-likelihood; similar objectives based on max-margin
are also possible \citep{liang2015semantics}.  It can be helpful to also add an
$L_1$ regularization term $\lambda \|\theta\|_1$,
which encourages feature weights to be zero, which produces a more compact
model that generalizes better \citep{berant2014paraphrasing}.
In addition, one can use AdaGrad \citep{duchi10adagrad},
which maintains a separate step size for each feature.
This can improve stability and convergence.

\section{Datasets and Results}
\label{sec:datasets}

In a strong sense, datasets are the main driver of progress for statistical approaches.
We will now survey some of the existing datasets, describe their
properties, and discuss the state-of-the-art.

The Geo880 dataset \citep{zelle96geoquery}
drove nearly a decade of semantic parsing research.
This dataset consists of 880 questions and database queries
about US geography (e.g., \nl{what is the highest point in the largest state?}).
The utterances are compositional,
but the language is clean and the domain is limited. %
On this dataset, learning from logical forms \citep{kwiatkowski2013scaling} and
answers \citep{liang11dcs} both achieve around 90\%

The ATIS-3 dataset \citep{zettlemoyer07relaxed}
consists of 5418 utterances paired with logical forms
(e.g., \nl{show me information on american airlines from fort worth texas to philadelphia}).
The utterances contain more disfluencies and flexible word order compared with Geo880,
\new{but they are logically simpler.
As a result, slot filling methods have been a successful paradigm in the spoken
language understanding community for this domain since the 1990s.
The best reported result is based on semantic parsing and obtains 84.6\%
}

The Regexp824 dataset \citep{kushman2013regex}
consists of 824 natural language and regular expression pairs
(e.g., \nl{three letter word starting with 'X'}).
The main challenge here is that there are many logically equivalent regular
expressions, some aligning better to the natural language than others.
\citep{kushman2013regex} uses semantic unification to test for logical form equivalence
and obtains 65.6\%

The Free917 dataset \citep{cai2013large} consists of 917 examples of
question-logical form pairs that can be answered via Freebase, e.g.
\nl{how many works did mozart dedicate to joseph haydn?}
The questions are \new{logically less complex than those in} the semantic parsing datasets above,
but introduces the new challenge of scaling up to many more predicates
(but is in practice manageable by assuming perfect named entity resolution
and leveraging the strong type constraints in Freebase).
The state-of-the-art accuracy is 68\%

WebQuestions \citep{berant2013freebase} is another dataset on Freebase
consisting of 5810 question-answer pairs (no logical forms)
such as \nl{what do australia call their money?}
Like Free917, the questions are not very compositional,
but unlike Free917,
they are real questions asked by people on the Web independent from Freebase,
so they are more realistic and more varied.
Because the answers are required to come from a single Freebase page,
a noticeable fraction of the answers are imperfect.
The current state-of-the-art is 52.5\%

The goal of WikiTableQuestions \citep{pasupat2015compositional}
is to extend question answering beyond Freebase to HTML tables on Wikipedia,
which are semi-structured.
The dataset consists of 22033 question-table-answer triples (e.g., \nl{how many runners took 2 minutes at the most to run 1500 meters?}), where
each question can be answered by aggregating information across the table.
At test time, we are given new tables, so methods must learn how to generalize
to new predicates.  The result on this new dataset is 37.1\%

\citet{wang2015overnight} proposed a new recipe for quickly using crowdsourcing
to generate new compositional semantic parsing datasets
consisting of question-logical form pairs.
Using this recipe,
they created eight new datasets in small domains consisting of 12602 total
question-answer pairs,
and achieved an average accuracy on across datasets of 58.8\%

\citep{chen11navigate} introduced a dataset of 706 navigation instructions
(e.g., \nl{facing the lamp go until you reach a chair}) in a simple grid world.
Each instruction sequence contains multiple sentences
with various imperative and context-dependent constructions not found in previous datasets.
\citet{artzi2013weakly} obtained 65.3\%

\section{Discussion}
\label{sec:discussion}

We have presented a semantic parsing framework for the problem of natural
language understanding.  Going forward, the two big questions are
(i) how to represent the semantics of language and
(ii) what supervision to use to learn the semantics from.

\myparagraph{Alternative semantic representations}
One of the main difficulties with semantic parsing is the divergence between the structure
of the natural language and the logical forms---purely compositional semantics
will not work.  This has led to some efforts to introduce an intermediate layer
between utterances and logical forms.
One idea is to use general paraphrasing models to map input utterances to the
``canonical utterances'' of logical forms
\citep{berant2014paraphrasing,wang2015overnight}.
This reduces semantic parsing to a text-only problem for which there is much more data and resources.

\new{One could also use domain-general logical forms that capture the
basic predicate-argument structures of sentences \citep{kwiatkowski2013scaling}.
Abstract meaning representation (AMR) \citep{banarescu2013amr} is one
popular representation backed by an extension linguistic annotation effort.
Multiple AMR parsers have been developed, including one based on CCG \citep{artzi2015broad}.
While these representations offer broad coverage,
solving downstream understanding tasks still require additional work
(e.g., inferring that \nl{population of X} is synonymous with
\nl{number of people living in X}).
In contrast, \emph{executable} semantic parsing operates
on the full pipeline from utterance to task output,
but compromises on domain-generality.
This is partially inevitable, as any understanding task requires some
domain-specific knowledge or grounding.
Designing the best general representation that supports
many downstream tasks remains an open challenge.
}

\myparagraph{Alternative supervision}
Much of the progress in semantic parsing has been due to being able to learn
from weaker supervision.
In the framework we presented,
this supervision are the desired actions $y$ (e.g., answers to questions).
One can use a large corpus of text to exploit even weaker supervision
\citep{krishnamurthy2012weakly,reddy2014large}.
More generally, one can think about language interpretation in a reinforcement learning
setting \citep{branavan09reinforcement},
where an agent who presented with an utterance in some context performs some action,
and receives a corresponding reward signal.
This framework highlights the importance of
context-dependence in language interpretation \citep{zettlemoyer09context,artzi2013weakly}.

\new{Due to their empirical success, there has been a recent surge of interest in using recurrent neural networks
and their extensions for solving NLP tasks such as machine translation %
and question answering \citep{yih2015stagg,weston2015memory}.
These methods share the same spirit of end-to-end utterance-to-behavior
learning as executable semantic parsing,
but they do not explicitly separate parsing from execution.
This makes them architecturally simpler than semantic parsers,
but they are more data hungry and it is unclear
whether they can learn to perform complex logical reasoning in a generalizable way.
Nonetheless, it is quite likely that these methods will play an important role
in the story of language understanding.
}

\myparagraph{Outlook}
This is an exciting time for semantic parsing and natural language
understanding.  As natural language interfaces (e.g., Siri)
become more ubiquitous, the demand for deep understanding will continue to grow.  At the
same time, there is a fresh wave of ambition pushing the limits of what can be
machine learnable.  The confluence of these two factors
will likely generate both end-user impact as well as new
insights into the nature of language and learning.

\small
\bibliographystyle{abbrv}
\bibliography{all}

\end{document}